\documentclass[5p, twocolumn]{elsarticle}
\pdfoutput=1
\usepackage[table,xcdraw]{xcolor}
\usepackage{lineno}
\usepackage{textcomp}
\usepackage[utf8]{inputenc}
\usepackage[T1]{fontenc}
\usepackage{url}
\usepackage{parskip} 
\usepackage{booktabs}
\usepackage{tikz,pgfplots}
\usepackage{color}
\usepackage{verbatim}
\usepackage{multirow}
\usepackage{lipsum}
\usepackage{subcaption}
\usepackage{booktabs}
\usepackage{afterpage}
\usepackage{soul}

\modulolinenumbers[20]

\begin{document}
\begin{frontmatter}

\title{Location Sensitive Deep Convolutional Neural Networks for Segmentation of White Matter Hyperintensities}

\author[uni,hospital]{Mohsen Ghafoorian\footnotemark[1]}
\author[hospital]{Nico Karssemeijer}
\author[uni]{Tom Heskes}
\author[neuro]{Inge W.M. van Uden}
\author[hospital]{Clara I. Sanchez}
\author[hospital]{Geert Litjens}
\author[neuro]{Frank-Erik de Leeuw}
\author[hospital]{Bram van Ginneken}
\author[uni]{Elena Marchiori}
\author[hospital]{Bram Platel}

\address[uni]{Institute for Computing and Information Sciences, Radboud University, Nijmegen, The Netherlands}
\address[hospital]{Diagnostic Image Analysis Group, Department of Radiology and Nuclear Medicine, Radboud University Medical Center, Nijmegen, The Netherlands}
\address[neuro]{Donders Institute for Brain, Cognition and Behaviour, Department of Neurology, Radboud University Medical Center, Nijmegen, The Netherlands}
\begin{abstract}
The anatomical location of imaging features is of crucial importance for accurate diagnosis in many medical tasks. Convolutional neural networks (CNN) have had huge successes in computer vision, but they lack the natural ability to incorporate the anatomical location in their decision making process, hindering success in some medical image analysis tasks.\\
In this paper, to integrate the anatomical location information into the network, we propose several deep CNN architectures that consider multi-scale patches or take explicit location features while training. We apply and compare the proposed architectures for segmentation of white matter hyperintensities in brain MR images on a large dataset. As a result, we observe that the CNNs that incorporate location information substantially outperform a conventional segmentation method with hand-crafted features as well as CNNs that do not integrate location information. On a test set of 46 scans, the best configuration of our networks obtained a Dice score of 0.791, compared to 0.797 for an independent human observer. Performance levels of the machine and the independent human observer were not statistically significantly different (p-value=0.17). 
\end{abstract}


\end{frontmatter}
 \biboptions{sort&compress}
%\linenumbers
%*************************************************************************************************************************************************************************************
\section{Introduction}
\footnotetext[1]{Email: mohsen.ghafoorian@radboudumc.nl, phone: +31 243655793, fax: +31 24 3652728}
White matter hyperintensities (WMH), also known as leukoaraiosis or white matter lesions are a common finding on brain MR images of patients diagnosed with small vessel disease (SVD) 
\citep{Norden11c}, multiple sclerosis \citep{schoonheim2014sex}, Parkinsonism \citep{marshall2006white}, stroke \citep{weinstein2013brain}, Alzheimer’s disease \citep{hirono2000impact} and Dementia \citep{smith2000white}.
WMHs often represent areas of demyelination found in the white matter of the brain, but they can also be caused by other mechanisms such as edema. WMHs are best observable in fluid-attenuated inversion recovery (FLAIR) MR images, as high value signals \citep{wardlaw2013neuroimaging}. The prevalence of WMHs among SVD patients has been reported to reach up to 95\% depending on the population studied and the imaging technique used \citep{de2001prevalence}. Studies have reported a relationship between WMH severity and other neurological disturbances and symptoms including 
cognitive decline \citep{de2000cerebral,au2006association}, gait dysfunction \citep{whitman2001prospective}, hypertension \citep{firbank2007brain} as well as depression \citep{herrmann2008white} and mood disturbances \citep{van2015white}. It has been shown that using a more accurate WMH volumetric assessment, a better association with clinical measures of physical performance and cognition is achieved \citep{van2006impact}. 

Accurate quantification of WMHs in terms of total volume and distribution is believed to be of clinical importance for prognosis, tracking of disease progression and assessment of the treatment effectiveness \citep{polman2005diagnostic}. However, manual segmentation of WMHs is a laborious time consuming task that makes it infeasible for larger datasets and in clinical practice. Furthermore, manual segmentation is subject to considerable inter- and intra-rater variability \citep{grimaud1996quantification}.

In the last decade, many automated and semi-automated algorithms have been proposed that can be classified into two general categories. Some methods use supervised machine learning algorithms, often using hand-crafted features \citep{anbeek2004probabilistic,lao2008computer,herskovits2008automated,simoes2013automatic,ithapu2014extracting,ghafoorian2015small,kloppel2011comparison, 
zijdenbos2002automatic, dyrby2008segmentation, geremia2011spatial,ghafoorian2016aut} or more recently with learned representations \citep{vijverberg2016single,brosch2016deep,brosch2015deep,ghafoorian2016non, 
kamnitsas2016efficient}. This is while other methods use unsupervised approaches \citep{van2001automated,shi2013automated,khademi2012robust,admiraal2005fully,de2009white,jain2015automatic, 
shiee2010topology,schmidt2012automated} to cluster WMHs as outliers or model them with additional classes. Although a multitude of approaches has been suggested for this problem, a truly reliable fully automated method that performs as good as human readers has not been identified \citep{caligiuri2015automatic,garcia2013review}.

Deep neural networks \citep{lecun2015deep, schmidhuber2015deep} are biologically plausible learning structures, inspired by early neuroscience-related work \citep{hubel1962receptive, 
fukushima1980neocognitron} and have so far claimed human level or super-human performances in several different domains \citep{cirecsan2012multi, he2015delving, taigman2014deepface, 
ciresan2012deep, cirecsan2013multi}. Convolutional neural networks (CNN) \citep{lecun1998gradient}, perhaps the most popular form of deep neural networks, have attracted enormous attention from the computer vision community since Alex Krizhevsky's network \citep{krizhevsky2012imagenet} won the Imagenet competition \citep{deng2009imagenet} by a large margin. Although the initial focus of CNN methods was concentrated on image classification, soon the framework was extended to cover segmentation as well. A natural way to apply CNNs to segmentation tasks is to train a network in a sliding-window setup to predict the label of each pixel/voxel considering a local neighborhood, which is usually referred to as a patch \citep{ciresan2012deep,
farabet2013learning,gupta2014learning,hariharan2014simultaneous}. Later fully convolutional neural networks were proposed to computationally optimize the segmentation process \citep{long2014fully,ronneberger2015u}.

Deep neural networks have recently been widely used in many medical image analysis domains including lesion detection, image segmentation, shape modeling and image registration \citep{greenspan2016guest}. In particular on neuroimaging, several studies are proposed using CNNs for brain extraction \citep{kleesiek2016deep}, tissue and anatomical region segmentation \citep{zhang2015deep,moeskops2016automatic,milletari2016hough,chen2016voxresnet,nie2016fully,shakeri2016sub}, tumor segmentation \citep{pereira2016brain,havaei2016brain,havaei2016hemis,zhao2016multiscale}, microbleed detection \cite{dou2015automatic,dou2016automatic}, lacune detection \cite{ghafooriandeep2016}, and brain lesion segmentation \citep{brosch2016deep,brosch2015deep,kamnitsas2016efficient,ghafoorian2016non}.

\begin{figure}[t]
\centering
\centerline
{
	\includegraphics[width=6cm]{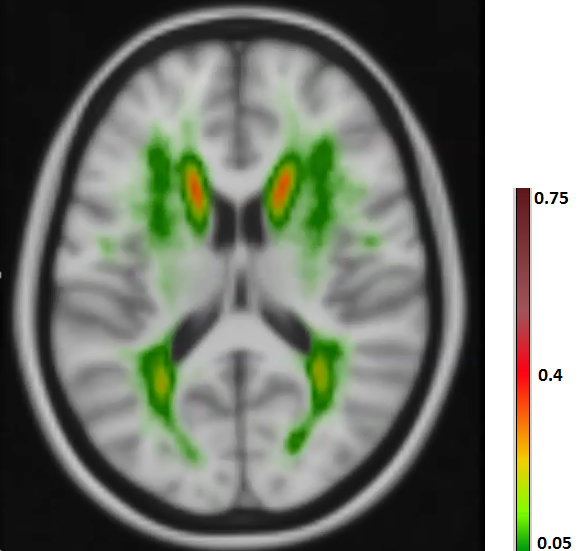}
}
\caption{A pattern is observable in WMHs occurrence probability map.}
\label{fig:LesionDistribution}
\end{figure}

In many bio-medical segmentation applications, including the segmentation of WMHs \citep{ghafoorian2015small, caligiuri2015automatic,garcia2013review, kamber1995model, anbeek2004probabilistic}, anatomical location information plays an important role for an accurate classification of voxels (see Figure \ref{fig:LesionDistribution}). In contrast, in commonly used segmentation benchmarks in the computer vision community, such as general scene labeling and crowd segmentation, it is normally not a valid assumption to consider pixel/voxel spatial location as an important piece of information. Integration of explicit location features has been recently tried on scoring of coronary calcium in cardiac CT \cite{wolterink2015automatic,Wolterink2016123}. However, we investigate the different possible locations to add the features and show it to be more effective to add the features in a different location compared to the one proposed in the mentioned study.

In this study, we train a number of CNNs to build systems for an accurate fully-automated segmentation of WMHs. We train, validate and evaluate our networks with a large dataset of more than 500 patients, that enables us to learn optimal values for millions of weights in our deep networks. In order to feed the CNN with location information, it is possible to incorporate multi-scale patches or add an explicit set of spatial features to the network. We evaluate and compare three different strategies and network architectures for providing the networks with more context/spatial location information. Experimental results suggest not only our best performing network outperforms a conventional segmentation method with hand-crafted features with a considerable margin, but also its performance does not significantly differ from an independent human observer. 

\begin{table*}[]
\centering
\caption{MR imaging protocol specification for the T1 and FLAIR modalities.}
\label{tab:MRProtocol}
\begin{tabular}{@{}lllll@{}}
\toprule
Modality & TR/TE/TI         & Flip angle & Voxel size     & Interslice gap \\ \midrule
T1       & 2250/3.68/850 ms 	&15\textdegree      		& 1.0$\times$1.0$\times$1.0 & 0              \\
FLAIR    & 9000/84/2200 ms  	&15\textdegree   	& 1.2$\times$1.0$\times$5.0 & 1 mm           \\ \bottomrule
\end{tabular}
\end{table*}

%&&&&&&&&&&&&&&&&&&&&&&&&&&&&&&&&&&&&&&&&&&&&&&&&&
%&&&&&&&&&&&&&&&&&&&&&&&&&&&&&&&&&&&&&&&&&&&&&&&&&
%*************************************************************************************************************************************************************************************
\section{Materials}

\subsection{Data}
The research presented in this paper uses data from a longitudinal study called the Radboud University Nijmegen Diffusion tensor and Magnetic resonance imaging Cohort (RUN DMC) \citep{Norden11c}. Baseline scanning was performed in 2006. The patients were rescanned in 2011/2012 and currently a third follow-up is being acquired.
\subsubsection{Subjects}
Subjects for the RUN DMC study were selected at baseline based on the following inclusion criteria \citep{Norden11c}: (a) aged between 50 and 85 years (b) cerebral SVD on neuroimaging (appearance of WMHs and/or lacunes). Exclusion criteria comprised: presence of (a) dementia (b) parkinson(-ism) (c) intracranial hemorrhage (d) life expectancy less than six months (e) intracranial space occupying lesion (f) (psychiatric) disease interfering with cognitive testing or follow-up (g) recent or current use of acetylcholine-esterase inhibitors, neuroleptic agents, L-dopa or dopa-a(nta)gonists (h) non-SVD related WMH (e.g. MS) (i) prominent visual or hearing impairment (j) language barrier and (k) MRI contraindications. Based on these criteria, MRI scans of 503 patients were taken at baseline.
\subsubsection{Magnetic resonance imaging}
The machine used for the baseline was a single 1.5 Tesla scanner (Magnetom Sonata, Siements Medical Solution, Erlangen, Germany). Details of the imaging protocol are listed in Table \ref{tab:MRProtocol}.
\subsubsection{Reference annotations}
Reference annotations were created in a slice by slice manner by two experienced raters, manually contouring hyperintense lesions on FLAIR MRI that did not show corresponding cerebrospinal fluid like hypo-intense lesions on the T1 weighted image. Gliosis surrounding lacunes and territorial infarcts were not considered to be WMH related to SVD \citep{herve2005shape}. One of the observers (observer 1) manually annotated all of the cases. 50 of these 503 images were selected at random and were annotated also by another human observer (observer 2). 
\begin{figure}[t]
\makebox[\linewidth]
{
	\begin{subfigure}[b]{.15\textwidth}
	\centering
	\includegraphics[height=2.8cm]{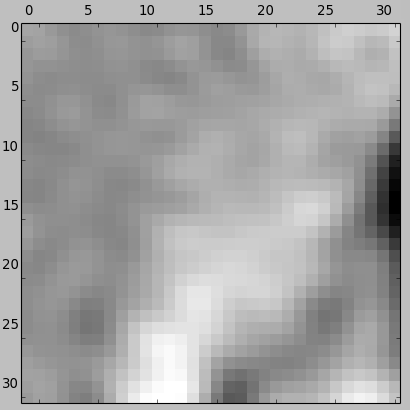} 
	\end{subfigure}

	\begin{subfigure}[b]{.15\textwidth}
	\centering
	\includegraphics[height=2.8cm]{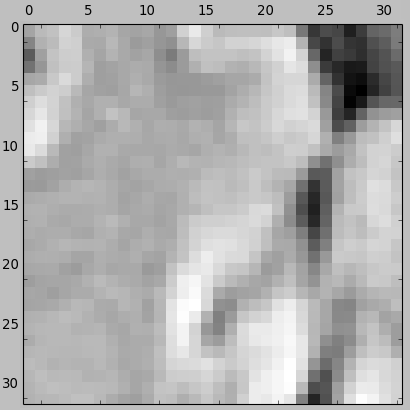}
	\end{subfigure}

	\begin{subfigure}[b]{.15\textwidth}
	\centering
	\includegraphics[height=2.8cm]{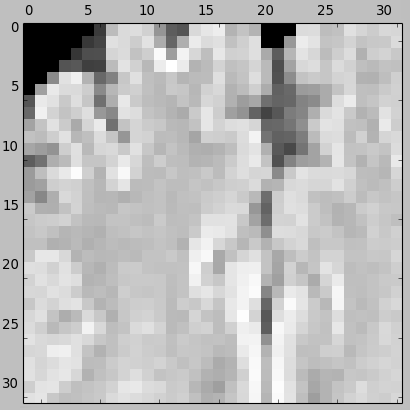}
	\end{subfigure}
}
\makebox[\linewidth]
{
	\begin{subfigure}[b]{.15\textwidth}
	\centering
	\includegraphics[height=2.8cm]{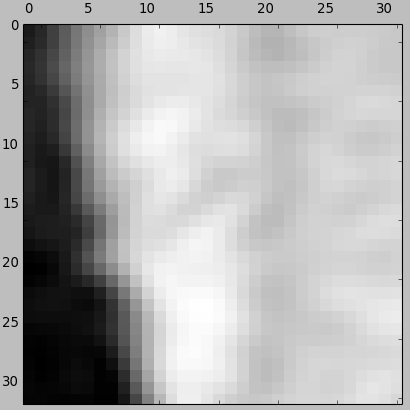} 
	\end{subfigure}

	\begin{subfigure}[b]{.15\textwidth}
	\centering
	\includegraphics[height=2.8cm]{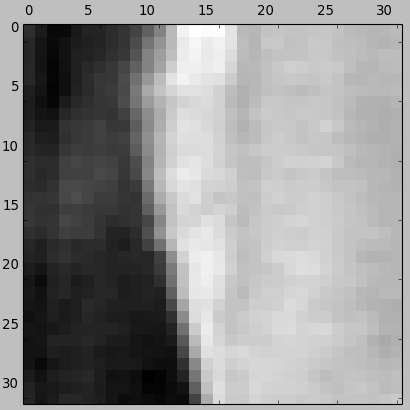}
	\end{subfigure}

	\begin{subfigure}[b]{.15\textwidth}
	\centering
	\includegraphics[height=2.8cm]{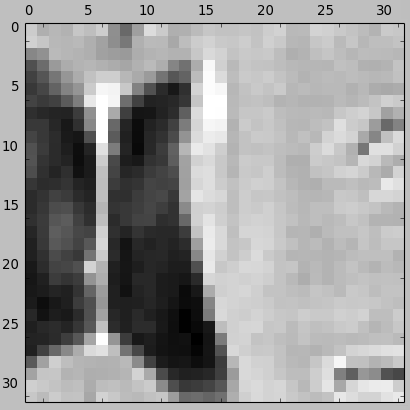}
	\end{subfigure}
}
\caption{An example of negative (top row) and positive (bottom row) samples in three scales (from left to right) 32$\times$32, 64$\times$64 and 128$\times$128 on the FLAIR image. The two larger scales are down sampled to 32$\times$32.}
\label{fig:samples}
\end{figure}

\subsection{Preprocessing}
Before supplying the data to our networks, we first pre-processed the data with the following four steps:
\subsubsection{Multi-modal registration}
Due to possible movement of patients during scanning, the image coordinates of the T1 and FLAIR modalities might not represent the same location. Thus we transformed 
the T1 image to align with the FLAIR image in the native space using FSL-FLIRT \citep{jenkinson2001global} implementation of rigid registration with trilinear interpolation and mutual information 
optimization criteria. Also to obtain a mapping between patient space and an atlas space, all subjects were non-linearly registered to the ICBM152 atlas \citep{mazziotta2001four} using FSL-FNIRT 
\citep{jenkinson2012fsl}. 

\subsubsection{Brain extraction}
In order to extract the brain and exclude other structures, such as skull, eyes, etc., we apply FSL-BET \citep{smith2002fast} on T1 images, because this modality has the highest resolution. The resulting mask is then transformed using registration transformation and is applied to the FLAIR images.
\subsubsection{Bias field correction}
Bias field correction is another necessary step due to magnetic field inhomogeneity. We apply FSL-FAST \citep{zhang2001segmentation}, which uses a hidden Markov random field and an associated expectation-maximization algorithm to correct for spatial intensity variations caused by RF inhomogeneities.
\subsubsection{Intensity normalization}
Apart from intensity variations caused by the bias field, intensities can also vary between patients. Thus we normalize the intensities per patient to be within the range of [0, 1].

\subsection{Training, validation and test sets}
From the 503 RUN DMC cases, we removed a number of cases that were extremely noisy or had failed in some of the preprocessing steps including brain extraction and registration, which left us with 420 out of 453 cases with single annotations and 46 cases out of 50 with double annotations. From 420 cases annotated by one human observer, we select 378 cases for training the model and the remaining 42 cases for validation and parameter tuning purposes. We use the 46 cases that were annotated by both human observers as independent test set. All the four cases that we left out from the test set were because of presence of severe noise as a result of head movement during image acquisition.\\
Medical datasets usually suffer from the fact that pathological observations are significantly less frequent compared to healthy observations, which also holds for our dataset. Given this, a simple uniform sampling may cause serious problems for the learning process \citep{pastor2013f}, as a classifier that labels all of the samples as normal, would achieve a high accuracy. To handle this, we undersample the negative samples to create a balanced dataset. We randomly select 50\% of positive and select an equal number of negative samples from normal voxels of all cases. This sampling procedure resulted in datasets consisting of 3.88 million and 430 thousand samples for training and validation sets respectively. 
%&&&&&&&&&&&&&&&&&&&&&&&&&&&&&&&&&&&&&&&&&&&&&&&&&
\begin{figure*}[t]
\centering
\centerline
{
	\includegraphics[width=6in]{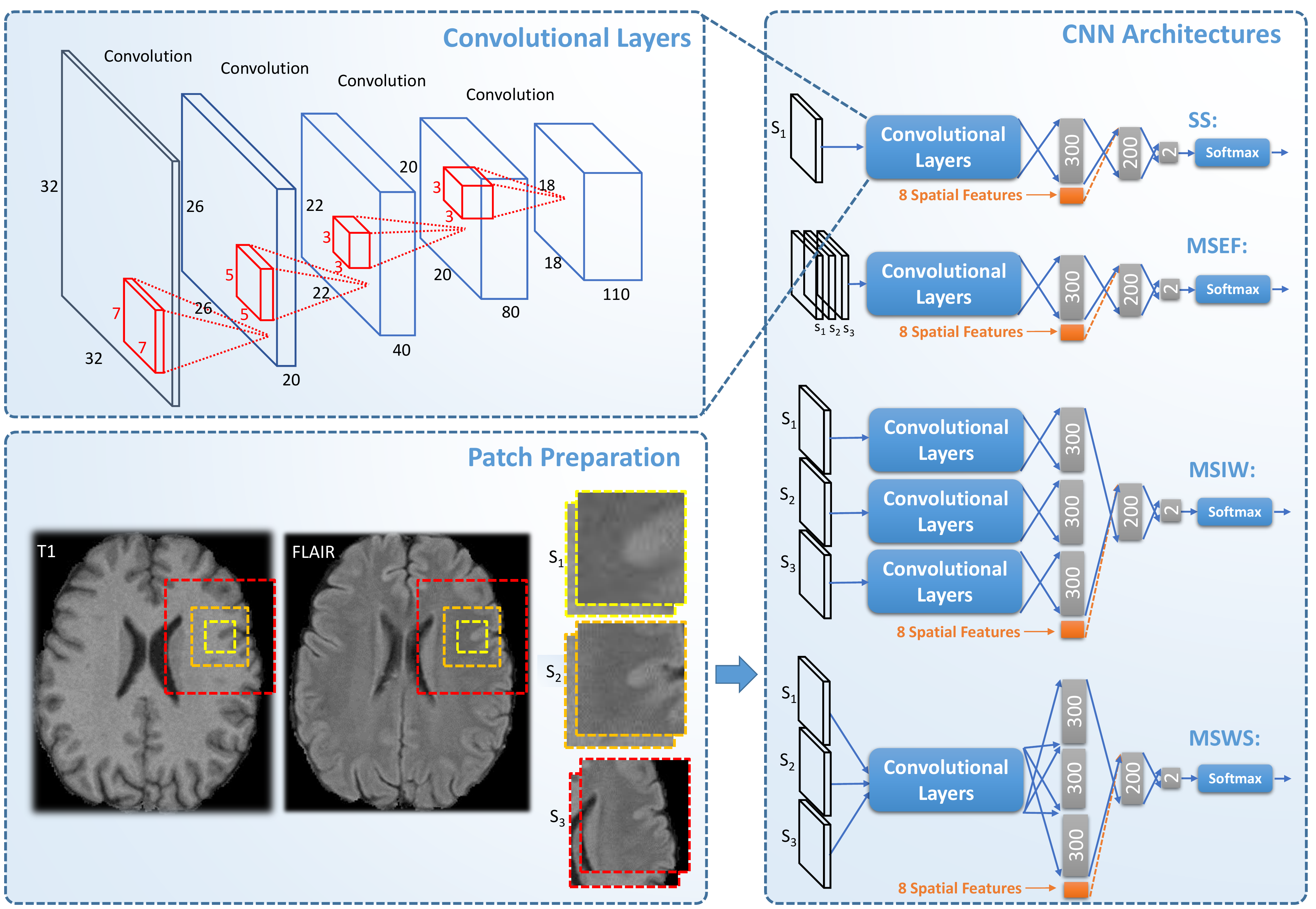}
}
\caption{Patch preparation process and different proposed CNN architectures.}
\label{fig:networks}
\end{figure*}
%&&&&&&&&&&&&&&&&&&&&&&&&&&&&&&&&&&&&&&&&&&&&&&&&&

%*************************************************************************************************************************************************************************************
\section{Methods}
\subsection{Patch preparation}
From each voxel neighborhood, we extract patches with three different sizes: 32$\times$32, 64$\times$64 and 128$\times$128. To reduce the computational costs, we down sample the larger two scales to 32$\times$32. Resulting patches for this procedure are demonstrated in Figure \ref{fig:samples}, for a negative and a positive sample, obtained from a FLAIR image. We included these three patches for both the T1 and FLAIR modalities for each sample. This results in a set of patches in three scales s$_1$, s$_2$ and s$_3$, each consisting of two patches from T1 and FLAIR, as depicted in Figure \ref{fig:networks}.
\subsection{Network architectures}
\subsubsection{Single-scale (SS) model}
The simplest CNN model we applied to our dataset was a CNN trained on patches from a single scale (with patches of 32$\times$32). The top architecture in Figure \ref{fig:networks} shows the architecture of our single-scale deep CNN. This network, which is a basis for the other location sensitive architectures, consists of four convolutional layers that have 20, 40, 80 and 110 filters of size 7$\times$7, 5$\times$5, 3$\times$3, 3$\times$3 respectively. We do not use pooling since it results in a shift-invariance property \citep{scherer2010evaluation}, which is not desired in segmentation tasks. Then we apply three layers of fully connected neurons of size 300, 200 and 2. Finally the resulting responses are turned into probability values using a softmax classifier.
\subsubsection{Multi-scale early fusion (MSEF)}
In many cases, it is impossible to correctly classify a 32$\times$32 patch just from its appearance. For instance, only looking at the small scale positive patch in Figure \ref{fig:samples}, it is hard to distinguish it from cortex tissue. In contrast, given the two larger scale patches, it is fairly easy to identify it as WMH tissue near the ventricles. Furthermore there is a trade-off between context capturing and localization accuracy. Although more context information might be captured with a larger patch-size, the ability of the classifier to accurately localize the structure in the center of the patch is decreased \citep{ronneberger2015u}. This motivates a multi-scale approach that has the advantages of the smaller and larger size patches. A simple and intuitive way to train a multi-scale network is to accumulate the different scales as different channels of the input. This is possible since the larger scale patches were down sampled to 32$\times$32. The second top network in Figure \ref{fig:networks} illustrates this.
\subsubsection{Multi-scale late fusion with independent weights (MSIW)}
Another possibility to create a model with multi-scale patches is to train independent convolutional layers for each scale, fusing the representations of each scale and taking them into more fully connected layers. As can be observed in Figure \ref{fig:networks}, in this architecture each scale has its own fully connected layer. These are concatenated and fed into the joint fully connected layers. The main rationale behind giving each scale stream its own fully connected layer is that this incurs less weights compared to the approach that firsts merges the feature maps and then fully connects it to the first layer of neurons.
\subsubsection{Multi-scale late fusion with weight sharing (MSWS)}
The first convolutional layers of a CNN typically detect various forms of edges, corners and other basic structuring elements. Since we do not expect that these basic building blocks differ much among the different scale patches, a considerable number of filters might be very similar in the three separate convolutional layers learned for different scales. Thus a potentially efficient strategy to reduce the number of weights and consequently to reduce the overfitting, is to share the convolutional filters among the different scales. As illustrated in Figure \ref{fig:networks}, each of the scales from the different patches are separately passed through a unified set of convolutional layers and each get described with separate feature maps. These feature maps are then connected to separate fully connected layers and are merged later, similar to the MSIW approach.
\subsubsection{Integrating explicit spatial location features}
The main aim for considering patches at different scales is to let the network learn about the spatial location of the samples it is observing. Alternatively we can provide the network with such information, by adding explicit features describing the spatial location. One possible place to add the location information is the first fully connected layer after the convolutional layers. All the location features are normalized per case to be within the range of [0, 1]. As the response of other neurons in the same layer that the location features are integrated with might have a different scale, all the eight features are scaled with a coefficient $\alpha$ as a parameter of the method. We tuned the best value for $\alpha$ as a parameter by validation.
The possibility to add spatial location features is not restricted to the single-scale architecture. It is also feasible to integrate these features into the three possible architectures for multi-scale approaches. The orange parts in Figure \ref{fig:networks} illustrate this procedure.\\ There are eight features that we utilize to describe the spatial location: the $x$, $y$ and $z$-coordinates of the corresponding voxel in the MNI atlas space, in-plane distances from the left ventricle, right ventricle, brain cortex and midsagittal brain surface as well as the prior probability of WMH occurring in that location \citep{ghafoorian2015small}. 
\subsection{Training procedure}
For learning the network weights, we use the stochastic gradient descent algorithm \citep{bottou2010large}, with mini-batch size of 128 and a cross-entropy cost. We also utilize the RMSPROP algorithm \citep{dauphin2015rmsprop} to speed up the learning process by adaptively changing the learning rate for each parameter. The non-linearity applied to neurons is a rectified linear unit to prevent the vanishing gradient problem \citep{maas2013rectifier}. As random weight initialization is important to break the symmetry between the units in the same hidden layer \citep{bottou2012stochastic}, the initial weights are drawn at random using the Glorot method \citep{glorot2010understanding}. Since CNNs are complex architectures, they are prone to overfit the data very early. Therefore we use drop-out regularization \citep{srivastava2014dropout} with 0.3 probability on all fully connected layers of the networks. We pick the resulting network from an epoch with the highest validation $A_z$ as the final model.
\begin{table*}[]
\small
\centering
\caption{Performance comparison of different CNN architectures based on validation set A$_z$ and test set Dice score considering observer 1 and observer 2 as the reference standard.}
\label{tab:dices}
\begin{tabular}{l|lll|lll}
\hline
                                              & \multicolumn{3}{l|}{Without location features}                                                                                                                                                     & \multicolumn{3}{l}{With location features}                                                                                                                                                         \\ \hline
Method & \begin{tabular}[c]{@{}l@{}}Validation \\ set A$_z$\end{tabular} & \begin{tabular}[c]{@{}l@{}}Test set\\ Dice (obs1)\end{tabular} & \begin{tabular}[c]{@{}l@{}}Test set\\ Dice (obs2)\end{tabular}  &  \begin{tabular}[c]{@{}l@{}}Validation \\ set A$_z$\end{tabular} & \begin{tabular}[c]{@{}l@{}}Test set\\ Dice (obs1)\end{tabular} & \begin{tabular}[c]{@{}l@{}}Test set\\ Dice (obs2)\end{tabular} \\ \hline

SS  				& 0.9939  & 0.730	& 0.729  & 0.9972	& 0.783	& 0.778 	\\
MSEF 				& 0.9947  & 0.758	& 0.747  & 0.9966	& 0.783	& 0.771	\\
MSIW				& 0.9966  & 0.775	& 0.762  & 0.9972  	& 0.791	& 0.781	\\
MSWS				& 0.9965  & 0.773	& 0.759  & 0.9973  	& 0.791	& 0.780	\\

\hline
\end{tabular}
\end{table*}

\section{Experimental Evaluation}
For characterization of WMHs, several different methods have been proposed in this study, some of which only use patch appearance features, while others use multi-scale patches or explicit location features to the network or both. In order to obtain segmentations, we apply the trained networks to classify all the voxels inside the brain mask in a sliding window fashion. A comparison between the performance of the mentioned methods, together with a comparison to performance of an independent human observer and a conventional method with hand-crafted features would be insightful.\\
Integrating the location information into the first fully connected layer, as depicted in the architectures Figure \ref{fig:networks}, is only one of the possibilities. We can alternatively add the spatial location features to one layer before or after, i.e. to the responses from the last convolutional layer and to the second fully connected layer. To evaluate the relative performance of each possibility, we also train single-scale networks with the two other possibilities and compare them to each other. In order to provide information on how much effect the dataset size has on the performance of the trained network, we present and compare the results of a MSWS+Loc network trained with 100\%, 50\%, 25\%, 12.5\% and 6.25\% of the total training images. 

\subsection{Metrics}
The Dice similarity index, also known as the Dice score, is the most widely used measure for evaluating the agreement between different segmentation methods and their reference standard segmentations. \citep{garcia2013review,caligiuri2015automatic}. It is computed as \begin {equation} Dice = \frac{2\times TP }{FP+FN+2\times TP} \end{equation} where the value varies between 0 for complete disagreement, and 1 representing complete agreement between the reference standard and the evaluated segmentation. A Dice similarity index of 0.7 or higher is usually considered a good segmentation in the literature \citep{caligiuri2015automatic}. To create binary masks out of probability maps resulting from CNNs, we find an optimal value as a threshold that maximizes the overall Dice score on the validation set. The optimal thresholds are computed separately for each method. We also present test set receiver operating characteristic (ROC) curves and validation set area under the ROC curve ($A_z$). For computing each of these measures, we only consider the voxels inside the brain mask, to avoid taking easy voxels belonging to the background into account.\\ 
For the statistical significance test, we created a 100 boot-straps by sampling 46 instances with replacement. Then the Dice scores were computed on each bootstrap for each of the two compared methods. Empirical p-values were reported as the proportion of bootstraps where the Dice score for method B was higher than A, when the null-hypothesis to reject was ``method A is no better than B''. If no such bootstrap existed, the p-value$<$0.01 was reported, representing a significant difference.

\subsection{Conventional segmentation system}
In order to evaluate the relative performance of the proposed deep learning systems, we also train a conventional segmentation system, using hand-crafted features \citep{ghafoorian2015small}. The set of hand-crafted features consists of 22 features in total: intensity features including FLAIR and T1 intensities, second order derivative features including multi-scale Laplacian of Gaussian ($\sigma$=1,2,4 mm), multi-scale determinant of Hessian (t=1,2,4 mm), vesselness filter ($\sigma$=1 mm), a multi-scale annular filter (t=1,2,4 mm), FLAIR intensity mean and standard deviation in a 16$\times$16 neighborhood, as well as the same 8 location features that were used in the previous subsection. We use a random forest classifier with 50 subtrees to train the model.

\section{Experimental Results}
Table \ref{tab:dices} represents a comparison on validation set $A_z$ and test set Dice score, for each of the methods, once without and another time with addition of spatial location features, considering observer 1 as the reference standard. Table \ref{tab:compHand_CNN_Exps} compares the performance of the conventional segmentation method, our late fusion multi-scale architecture with weight sharing and location information (MSWS+Loc), and the two human observers on the independent test set, with each observer as the reference standard.\\ P-values were computed as a result of patient-level boot-strapping on the test set and are presented in Table \ref{tab:pvalues}.

%%%%%%%%%%%%%%%%%%%%%%%%%%%%%%%%%%%%%%%%%%5
\begin{table*}[]
\centering
\caption{A performance comparison between conventional method, MSWS+Loc architecture, and human observers.}
\label{tab:compHand_CNN_Exps}
\begin{tabular}{@{}llll@{}}
\toprule
Method                                    & Dice (obs1)	& Dice (obs2)	\\ \midrule
Conventional 			                  & 0.710    		& 0.685      		            \\
MSWS+Loc							& 0.791     		& 0.780      		            \\
observer 1             					& -           		& 0.797	    		                  \\
observer 2                                & 0.797       		& -           	                  \\ \bottomrule
\end{tabular}
\end{table*}
%%%%%%%%%%%%%%%%%%%%%%%%%%%%%%%%%%%%%%%%%%5

%%%%%%%%%%%%%%%%%%%%%%%%%%%%%%%%%%%%%%%%%%5
\begin{table*}[]
\centering
\caption{Statistical significance test for pairwise comparison of the methods Dice score. $p_{ij}$ indicates the p-value for the null hypothesis that method $i$ is better than method $j$.}
\label{tab:pvalues}
\begin{tabular}{@{}lccccc@{}}
\toprule
Method   & MSWS          & SS+Loc        & MSWS+Loc      & Ind. Obs.       \\ \midrule
SS        & \textless0.01 & \textless0.01 & \textless0.01 & \textless0.01 \\
MSWS    & -             & \textless0.01 & \textless0.01 & \textless0.01 \\
SS+Loc     & -             & -             & 0.23          & 0.15          \\
MSWS+Loc  & -             & -             & -             & 0.17          \\ \bottomrule
\end{tabular}
\end{table*}
%%%%%%%%%%%%%%%%%%%%%%%%%%%%%%%%%%%%%%%%%%5

Regarding the different options for integration of the location information in the network, Table \ref{fig:locToAddLocComp} compares the performance of these options on the validation and training sets. Adding the spatial location information to the first fully connected layer results in a significantly better Dice score compared to the other two possibilities (p-value $<$ 0.01). 

%%%%%%%%%%%%%%%%%%%%%%%%%%%%%%%%%%%%%%%%%%5
\begin{table}[t]
\centering
\caption{A performance comparison of the single-scale architecture with different possible locations to add the spatial location information. Abbreviations: last convolutional layer (LCL), first fully connected layer (FFCL), second fully connected layer (SFCL).}
\label{fig:locToAddLocComp}
\begin{tabular}{@{}cccc@{}}
\toprule
Method        & \begin{tabular}[c]{@{}l@{}}Validation set $A_z$\end{tabular} & \begin{tabular}[c]{@{}l@{}}Test set  Dice\end{tabular}         \\ \midrule
LCL    & 0.9964                                                        & 0.759        \\
FFCL  & 0.9971                                                        & 0.783 \\
SFCL &  0.9967                                                        & 0.771          \\ \bottomrule
\end{tabular}
\end{table}
%%%%%%%%%%%%%%%%%%%%%%%%%%%%%%%%%%%%%%%%%%5

%%%%%%%%%%%%%%%%%%%%%%%%%%%%%%%%%%%%%%%%%%5
\begin{figure*}[!t]
\makebox[\linewidth][c]
{
	\begin{subfigure}[b]{.45\textwidth}
	\centering
	\captionsetup{justification=centering,margin=0.4cm}
	{
	\includegraphics[height=6.5cm]{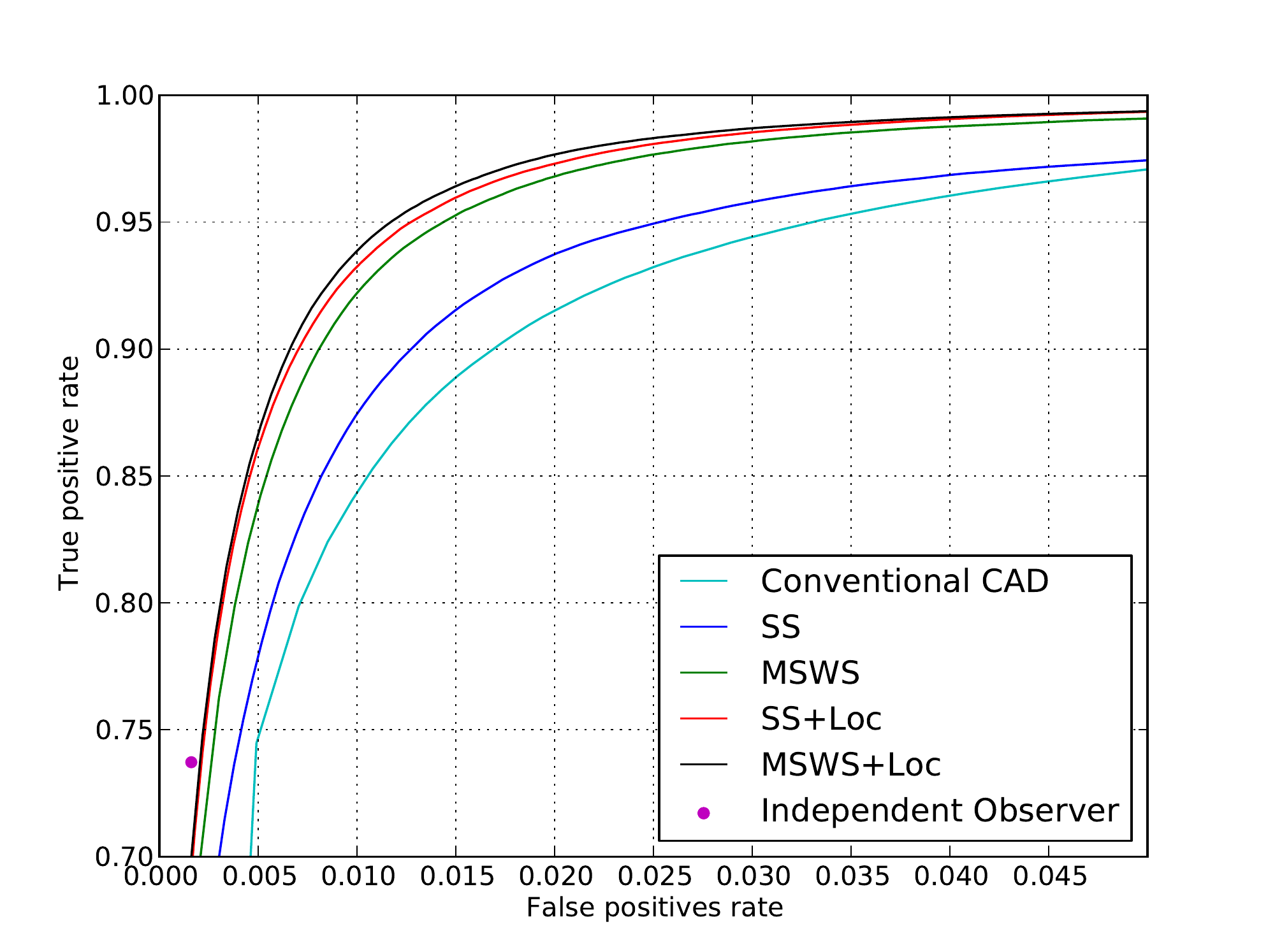} \caption{An ROC comparison of different CNN methods, a conventional segmentation method and independent human observer, considering observer 1 as the reference standard.} \label{fig:ROCr1}
	}
	\end{subfigure}
	\begin{subfigure}[b]{.45\textwidth}
	\centering
	\captionsetup{justification=centering,margin=0.1cm}
	{
	\includegraphics[height=6.5cm]{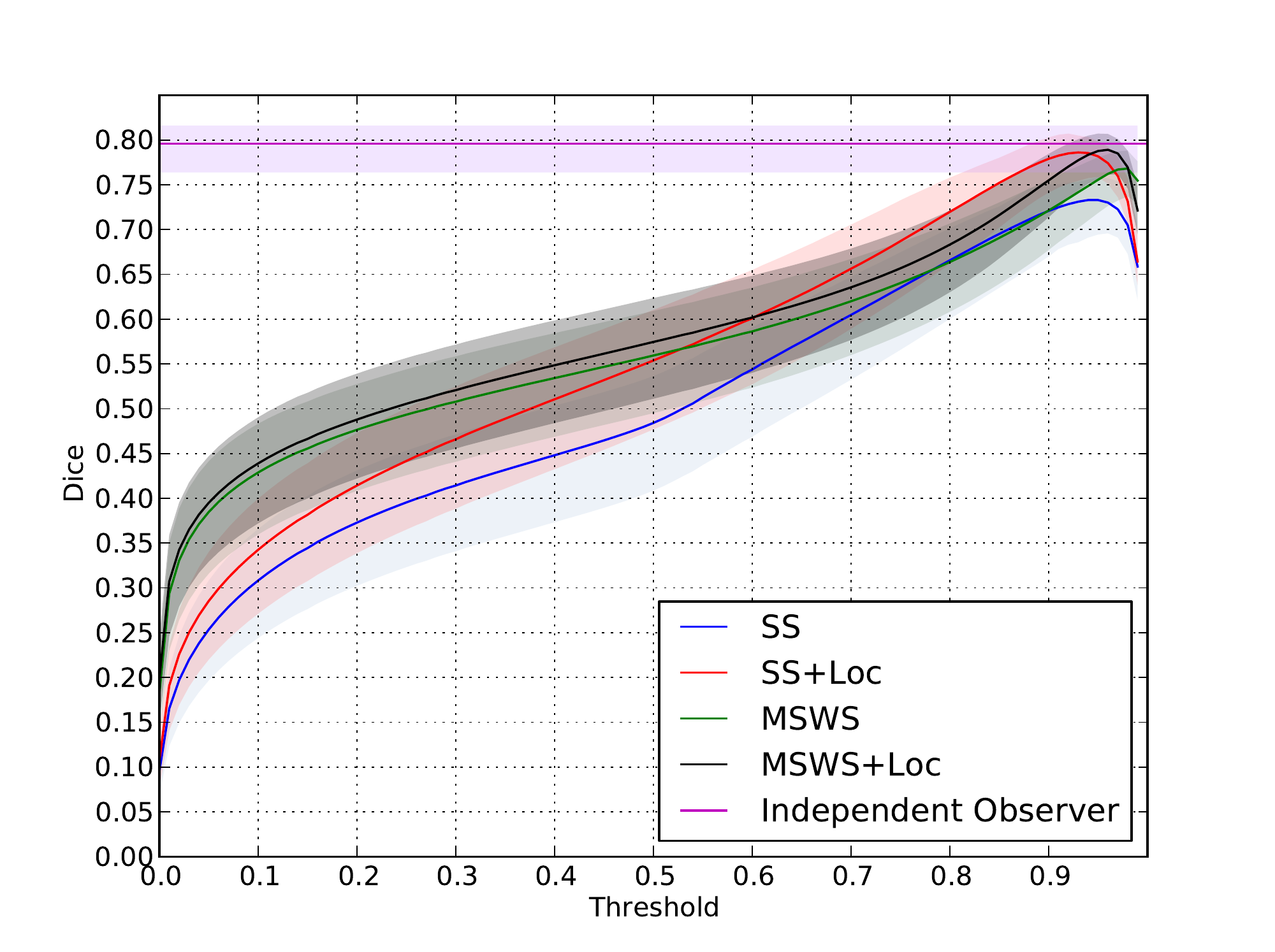} \caption{A comparison of different methods on Dice score as a function of binary masking threshold. The light shades around the curves indicate 95\% confidence intervals with bootstrapping on patients.} \label{fig:dice_r1}
	}
	\end{subfigure}
}
\caption{Integration of spatial location information fills the gap between performance of a normal CNN and human observer. }
\label{fig:ROCs}
\end{figure*}
%%%%%%%%%%%%%%%%%%%%%%%%%%%%%%%%%%%%%%%%%%5

\begin{figure}[t]
\centering
\centerline
{
	\includegraphics[width=9cm]{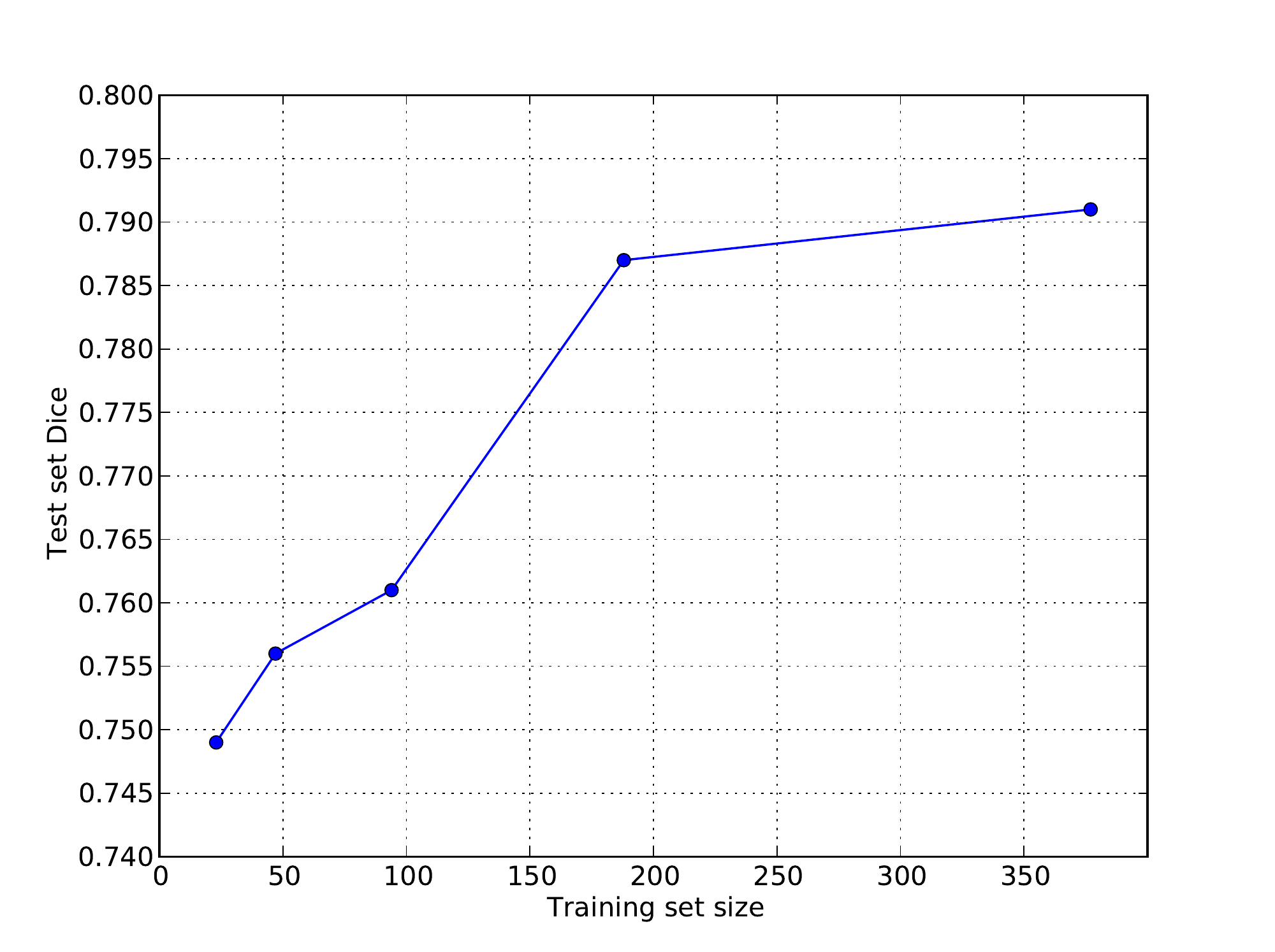}
}
\caption{Test Dice as a function of training set size.}
\label{fig:datasetSize}
\end{figure}

Figure \ref{fig:ROCr1} shows the ROC curves for some of the trained CNN architectures and compares them to the conventional segmentation method and the independent human observer. The ROC curves have been cut to show only low false positive rates that are of interest for practical use. In order to preserve readability of the figures, we only compare the most informative methods. Figure \ref{fig:dice_r1} shows the Dice similarity scores as a function of the binary masking threshold. It also compares them to the Dice similarity measure between the two human observers. 95\% confidence intervals are depicted for each curve, as a result of bootstrapping on patients. The effect of the training dataset size can be observed in Table \ref{tab:datasetSize} and Figure \ref{fig:datasetSize}. 

%%%%%%%%%%%%%%%%%%%%%%%%%%%%%%%%%%%%%%%%%%5

%%%%%%%%%%%%%%%%%%%%%%%%%%%%%%%%%%%%%%%%%%5

\section{Discussion}

% adding loc info or not? single-scale vs multi-scale
\subsection{Contribution of larger context and location information}
Comparing the performance of the SS and SS+Loc approaches, as presented in the first row of Table \ref{tab:dices}, a significant difference in Dice score is observable (p-value $<$ 0.01). This points us to the fact that a knowledge about where the input patch is located can substantially improve WMH segmentation quality of a CNN. A similar significant difference is observable when comparing performance measures of SS and MSWS methods (p-value $<$ 0.01). This implies that by using a multi-scale approach, a CNN can learn about context information quite well. Considering the better performance of SS+Loc compared to MSWS, we can infer that the learning of location and large scale context from multi-scale patches is not as good as adding explicit location information to the architecture.

% early fusion vs late fusion. independent vs weight sharing
\subsection{Early fusion vs. late fusion, independent weights vs. weight sharing}
As the experimental results suggest, among the different multi-scale fusion architectures, early fusion shows the least improvement over the single-scale approach. The related patch voxels of different scales, do not have a meaningful correspondence. Given the fact that the convolution operation in the first convolutional layer sums up the responses on each scale, we assume that the useful information provided by different scales is washed out too early in the network. In contrast, the two late fusion architectures show comparable good performance, however in general, since the late fusion architecture with weight sharing is a simpler model with less parameters to be learned, one might prefer to use this model.

%&&&&&&&&&&&&&&&&&&&&&&&&&&&&&&&&&&&&&&&&&&&&&&&&&&&&&&&&&&&&&&&&&&&&&&
\begin{figure*}[!t]
\makebox[\linewidth][c]
{
	\begin{subfigure}[b]{.2\textwidth}
		\centering
		\captionsetup{justification=centering,margin=0.1cm}
		{\includegraphics[height=8cm]{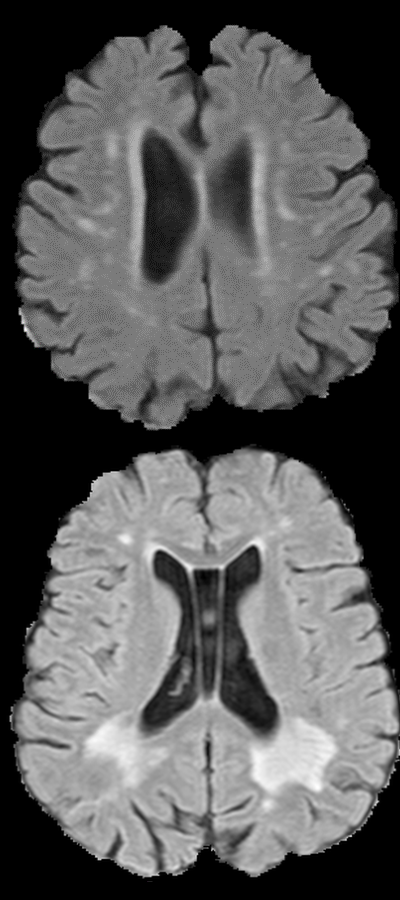}} 
		\caption{FLAIR images without annotations.}
	\label{fig:}
	\end{subfigure}%
	\begin{subfigure}[b]{.2\textwidth}
		\centering
		\captionsetup{justification=centering,margin=0.1cm}
		{\includegraphics[height=8cm]{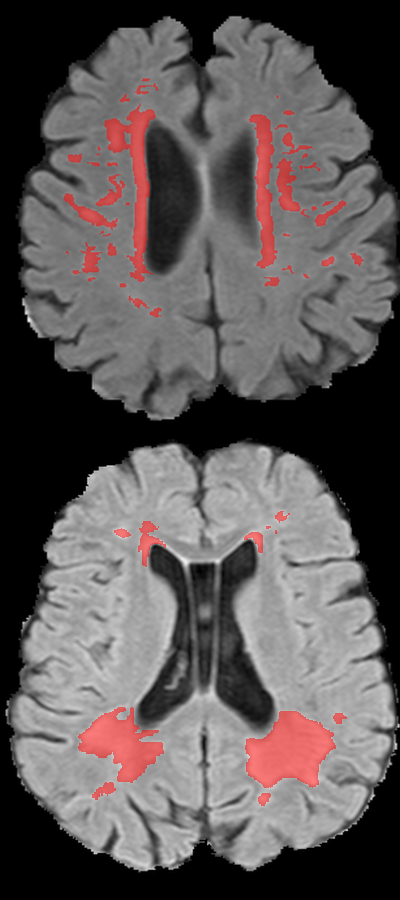}} 
		\caption{Segmentation by human observer 1.}
	\label{fig:}
	\end{subfigure}%
	\begin{subfigure}[b]{.2\textwidth}
		\centering
		\captionsetup{justification=centering,margin=0.1cm}
		{\includegraphics[height=8cm]{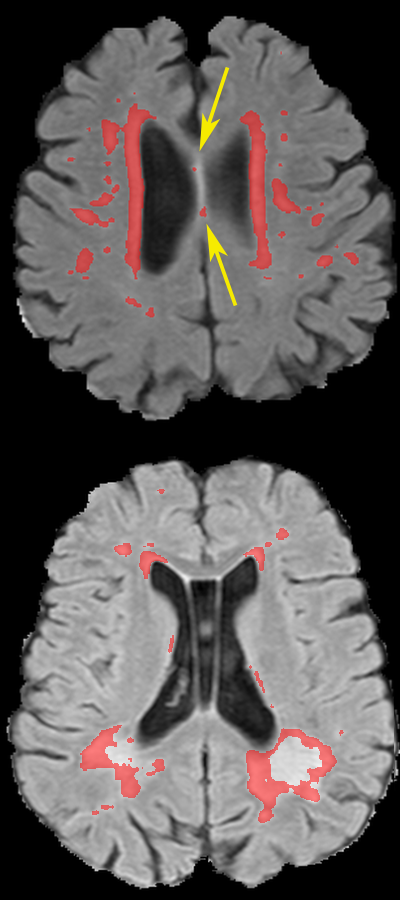}} 
		\caption{Segmentation by SS method.}
	\label{fig:}
	\end{subfigure}%
	\begin{subfigure}[b]{.2\textwidth}
		\centering
		\captionsetup{justification=centering,margin=0.1cm}
		{\includegraphics[height=8cm]{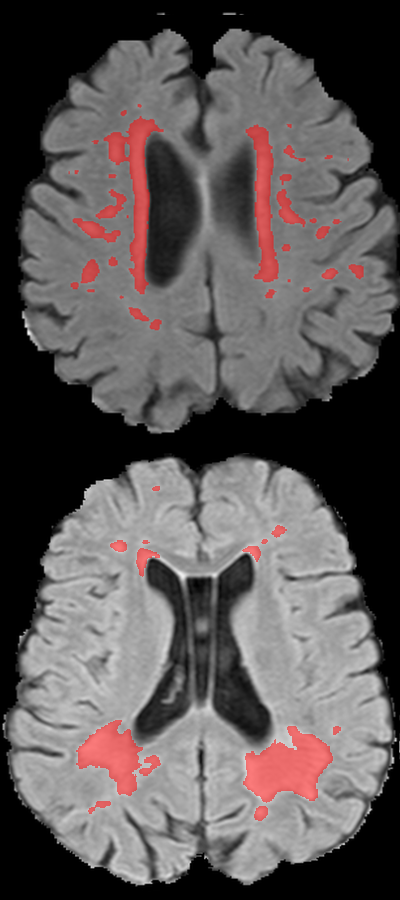}} 
		\caption{Segmentation by MSWS+Loc method.}
	\label{fig:}
	\end{subfigure}%
}
\caption{Two sample cases of segmentation improvement by adding location information to the network.}
\label{fig:SSvsMSWS}
\end{figure*}
%&&&&&&&&&&&&&&&&&&&&&&&&&&&&&&&&&&&&&&&&&&&&&&&&&&&&&&&&&&&&&&&&&&&&&&

\begin{table}[t]
\centering
\caption{Test Dice as a function of training set size.}
\label{tab:datasetSize}
\begin{tabular}{@{}lccccc@{}}
\toprule
Training set size & 23    & 47    & 94    & 189   & 378   \\ \midrule
Test set Dice     & 0.749 & 0.756 & 0.761 & 0.787 & 0.791 \\ \bottomrule
\end{tabular}
\end{table}

%&&&&&&&&&&&&&&&&&&&&&&&&&&&&&&&&&&&&&&&&&&&&&&&&&&&&&&&&&&&&&&&&&&&&&&
\begin{figure*}[!ht]
\makebox[\linewidth][c]
{
	\begin{subfigure}[b]{.2\textwidth}
		\centering
		\captionsetup{justification=centering,margin=0.1cm}
		{\includegraphics[height=4cm]{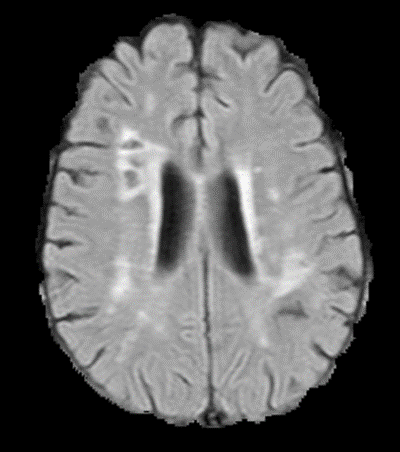}} 
		\caption{FLAIR images without annotations.}
	\label{fig:}
	\end{subfigure}%
	\begin{subfigure}[b]{.2\textwidth}
		\centering
		\captionsetup{justification=centering,margin=0.1cm}
		{\includegraphics[height=4cm]{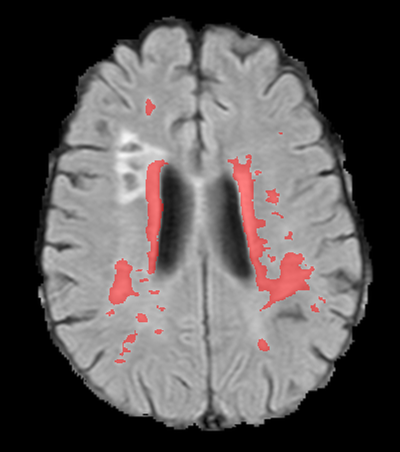}} 
		\caption{Segmentation by human observer 1.}
	\label{fig:}
	\end{subfigure}%
	\begin{subfigure}[b]{.2\textwidth}
		\centering
		\captionsetup{justification=centering,margin=0.1cm}
		{\includegraphics[height=4cm]{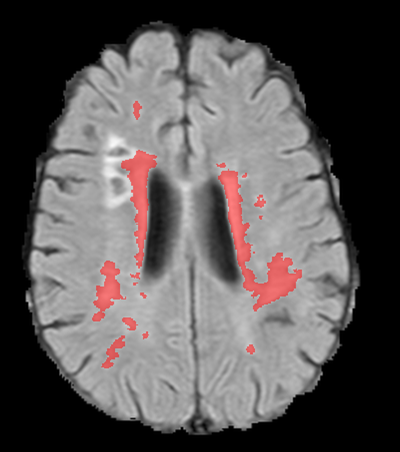}} 
		\caption{Segmentation by human observer 2.}
	\label{fig:}
	\end{subfigure}%
	\begin{subfigure}[b]{.2\textwidth}
		\centering
		\captionsetup{justification=centering,margin=0.1cm}
		{\includegraphics[height=4cm]{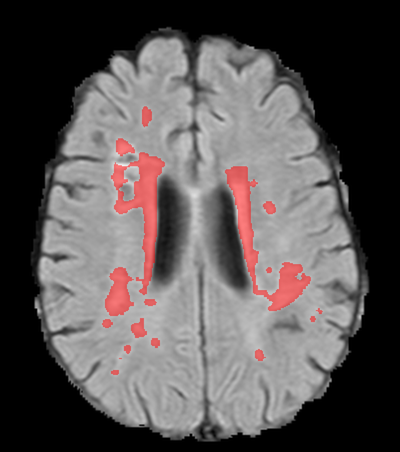}} 
		\caption{Segmentation by MSWS+Loc method.}
	\label{fig:}
	\end{subfigure}%
}
\caption{Gliosis around the lacunes is a prevalent type of false positive segmentation.}
\label{fig:FP}
\end{figure*}
%&&&&&&&&&&&&&&&&&&&&&&&&&&&&&&&&&&&&&&&&&&&&&&&&&&&&&&&&&&&&&&&&&&&&&&

%&&&&&&&&&&&&&&&&&&&&&&&&&&&&&&&&&&&&&&&&&&&&&&&&&&&&&&&&&&&&&&&&&&&&&&
\begin{figure*}[!ht]
\makebox[\linewidth][c]
{
	\begin{subfigure}[b]{.2\textwidth}
		\centering
		\captionsetup{justification=centering,margin=0.1cm}
		{\includegraphics[height=4cm]{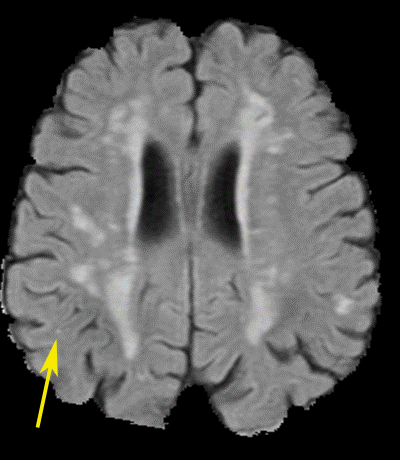}} 
		\caption{FLAIR image without annotations.}
	\label{fig:}
	\end{subfigure}%
	\begin{subfigure}[b]{.2\textwidth}
		\centering
		\captionsetup{justification=centering,margin=0.1cm}
		{\includegraphics[height=4cm]{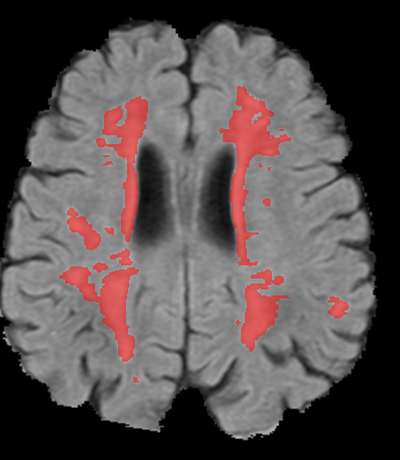}} 
		\caption{Segmentation by human observer 1.}
	\label{fig:}
	\end{subfigure}%
	\begin{subfigure}[b]{.2\textwidth}
		\centering
		\captionsetup{justification=centering,margin=0.1cm}
		{\includegraphics[height=4cm]{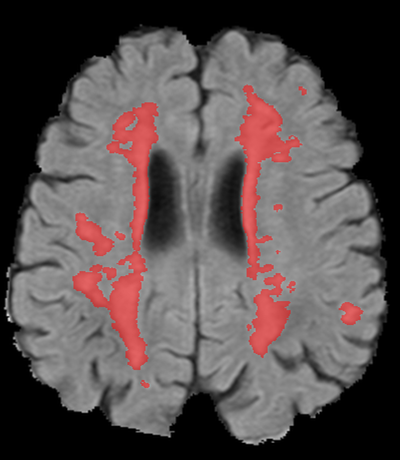}} 
		\caption{Segmentation by human observer 2.}
	\label{fig:}
	\end{subfigure}%
	\begin{subfigure}[b]{.2\textwidth}
		\centering
		\captionsetup{justification=centering,margin=0.1cm}
		{\includegraphics[height=4cm]{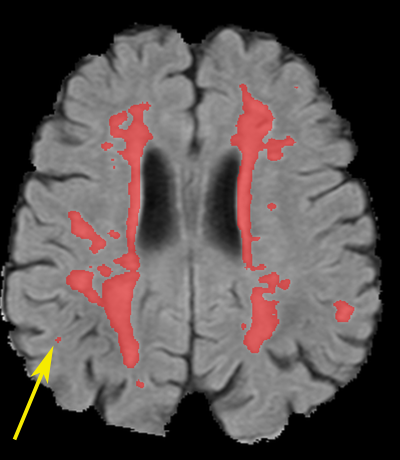}} 
		\caption{Segmentation by MSWS+Loc method.}
	\label{fig:}
	\end{subfigure}%
}
\caption{A sample case with a small lesion missed by the two human observers.}
\label{fig:missed}
\end{figure*}
%&&&&&&&&&&&&&&&&&&&&&&&&&&&&&&&&&&&&&&&&&&&&&&&&&&&&&&&&&&&&&&&&&&&&&&

%comparing the results to human expert and conventional CAD
\subsection{Comparison to human observer and a conventional method}
Shown by Table \ref{tab:compHand_CNN_Exps}, MSWS+Loc substantially outperforms a conventional segmentation method, with Dice score of 0.79 compared to 0.71 (p-value$<$0.01). Furthermore, the Dice score of MSWS+Loc method closely resembles the inter-observer variability, which implies that the segmentation provided by MSWS+Loc approach is as good as the two human observers. Also the statistical test does not show a significant advantage of the independent observer compared to this method (p-value = 0.17). 
%visual representation of the results

\subsection{A visual look into the results}
Figures \ref{fig:SSvsMSWS}-\ref{fig:missed} show some qualitative examples.  Figure \ref{fig:SSvsMSWS} contains two sample cases, where the location and larger context information leads to a better segmentation. As evident from the first sample, the single-scale CNN falsely segments an area on septum pellucidum, which also appears as hyperintense tissue. These false positives can be avoided by considering location information. A second sample shows improvements on FNs of the single-scale method.

Figure \ref{fig:FP} illustrates an instance of a prevalent class of false positives of the system, which are the hyperintense voxels around the lacunes. Since the model has not been trained on so many negative samples similar to this, the distinction between WMH and hyperintensities around lacunes is not well learned by the system. An obvious solution is to extensively include the lacunes surrounding voxels as negative samples in the training dataset.

As an example of missed lesions by human observers, Figure \ref{fig:missed} shows a small lesion on the right temporal lobe, missed by both human observers, where it is detected by MSWS+Loc method. Another sample of such missed lesions can be observed in the second sample of Figure \ref{fig:SSvsMSWS}, on the right hemisphere frontal lobe. Based on similar observations, we can assume that some of the false positives are possibly small lesions missed by one or both of the observers. Therefore there may be a chance that the real performance of the system is better than reported, but it would require more research to investigate this.

%A paragraph here comparing different possible locations to add the location information.
\subsection {Integration of location features}
For integration of explicit spatial location information into the CNN, there are several possibilities that were investigated in this study. The results as represented in Table \ref{tab:compHand_CNN_Exps}, suggest that adding the spatial location features to the first fully connected layer results in a significantly better performance. Adding them to around 35K features as the responses of the last convolutional layer, almost makes the eight location features insignificant among so many representation features. At the other extreme, although integrating the location features into the second fully connected layer does not suffer from this problem, but leaves less flexibility for the network to consider location features for the discrimination to be learned. The first fully connected layer seems to be the best option, where the appearance features provided by the last convolutional layer are already considerably reduced, and at same time the more fully connected layer provides more flexibility for an optimal discrimination.
% two-stage vs one-stage
\subsection{Two-stage vs. single-stage model}
As shown in the results, integrating location information into a CNN can play an important role in obtaining an accurate segmentation. We integrate the features while we train our network to learn the representations. Another approach is to perform this task in two stages; first training an independent network that learns the representations, and later training a second classifier that takes the output features of the first network, integrated with location or other external features (as followed in \cite{kooi2017large} for instance). The first approach, which is followed in this study, seems more reasonable as the set of learned filters without location information could differ from the optimal set of filters given the location information. The two-stage system lacks this information and might devote some of the filters for capturing of location that are redundant given the location features. 
%3D vs 2D patches
\subsection{2D vs. 3D patches}
In this research, we sample 2D patches from each of the two modalities (T1 and FLAIR), while one might argue that considering consecutive slices and sampling 3D patches from each image modality could provide useful information. Given the slice thickness of 5 mm with a 1 mm inter-slice gap in our dataset, the consecutive slices do not highly correspond to each other. Furthermore incorporation of 3D patches extensively increases the computational costs at both the training and the segmentation time. These motivated us to use 2D patches. In contrast, for datasets with isotropic or thin slice FLAIR images, 3D patches might be very useful.

\subsection{Fully convolutional segmentation network}
Fully convolutional networks replace the fully connected layers with 1$\times$1 convolutions that perform exactly the same functionality as the fully connected do, however implemented with convolutions \citep{long2014fully}. This would speed the segmentation up, since convolutions can get larger input images, make dense predictions for the whole input image and avoid repetitive computations. While we have trained our networks in a patch-based manner, it does not restrict us from reforming the fully connected layers of the trained network into convolutional layer counterparts at the segmentation time. The current implementation uses a patch-based segmentation, as we found it fast enough in the current experimental setup ($\sim$3 minutes for the multi-scale and $\sim$1.5 minutes for the single-scale architectures per case on a Titan X card).

\section{Conclusions}
In this study we showed that location information can have a significant added value when using CNNs for WMH segmentation. While for this task, making use of CNNs, not only a better performance compared to conventional segmentation method was achieved, we approached the performance level of an independent human observer with incorporation of location information.

\section{Acknowledgments}
This work was supported by a VIDI innovational grant from the Netherlands Organisation for Scientific Research (NWO, grant 016.126.351). The authors also would like to acknowledge Lucas J.B. van Oudheusden and Koen Vijverberg for their contributions to this study.

\section{References}
\bibliography{references}

\end{document}